\documentclass[runningheads]{llncs}
\usepackage[T1]{fontenc}
\usepackage{graphicx}
\usepackage{url}
\usepackage{hyperref}
\begin{document}
\title{Is text normalization relevant\\ for classifying medieval charters?}
\author{Florian Atzenhofer-Baumgartner\inst{1}\orcidID{0000-0001-8157-8629}\\ \and
Tamás Kovács\inst{1}\orcidID{0000-0002-3913-2946}}
\institute{Center for Information Modelling, University of Graz, Graz, Austria \\
\email{atzenhofer@acm.org}, \email{tamas.kovacs@uni-graz.at}}

\maketitle              
\begin{abstract}
This study examines the impact of historical text normalization on the classification of medieval charters, specifically focusing on document dating and locating. Using a data set of Middle High German charters from a digital archive, we evaluate various classifiers, including traditional and transformer-based models, with and without normalization. Our results indicate that the given normalization minimally improves locating tasks but reduces accuracy for dating, implying that original texts contain crucial features that normalization may obscure. We find that support vector machines and gradient boosting outperform other models, questioning the efficiency of transformers for this use case. Results suggest a selective approach to historical text normalization, emphasizing the significance of preserving some textual characteristics that are critical for classification tasks in document analysis.
\keywords{Document Classification  \and Historical Text Normalization\and Less-Resourced Languages \and Digital Diplomatics \and Digital Humanities}
\end{abstract}
\section{Introduction}
Charters are historical legal documents that are created and authenticated according to certain formal standards. They are crucial in confirming transactions and contracts between (multiple) parties, e.g., about land, property, privileges. These documents are invaluable for historical research, especially in studying the medieval and early modern period. Their importance is deeply connected to the core of human community life, which relies heavily on legal agreements~\cite{Ferraris_2014}. Charters have typically been preserved as original manuscripts, copies, or within (printed) scholarly editions.

The field of diplomatics is dedicated to critically examining charters, with a focus on affirming their authenticity. Doing so requires comparing many documents. Diplomatists thus face a significant challenge: efficiently sorting through vast amounts of these sources to find those most relevant to their research. A promising solution is found in machine-assisted text classification. Two tasks essential for historical research include dating and locating documents~\cite{He_Samara_Burgers_Schomaker_2016}. In the field of diplomatics, these tasks are crucial for approximating a charter's provenance, enabling the construction of so-called charter landscapes. These landscapes facilitate investigations into the evolution of legal norms and traditions, as well as linguistic features~\cite{Becker_Schallert_2021,Waldenberger_Dipper_Lemke_2021}.

While they are fundamental to historical research, these tasks are complicated by the inconsistency in text (preservation) traditions and the noise common in digitized forms of charters. Transcriptions of charters through automatic text recognition can intensify this trend~\cite{Torres_Aguilar_Jolivet_2023}. Since, beyond a certain threshold, there is no single best solution for historical text-normalization~\cite{Bollmann_2019}, and considering its immense effort, we must ask: is it justifiable considering its impact on text classification?

In light of this question, our work introduces several innovations. We utilize a recently introduced and curated charter data set of the later Middle High German (MHG) period, i.e., a historical predecessor of modern German, which has qualified as low-resource in the face of several NLP tasks~\cite{Nie_Schmid_Schütze_2023}. We also evaluate a range of classifiers, with an emphasis on balancing resource expenditure for a real-world task. Simultaneously, we assess how normalization affects classification performance. While this topic has recently received some attention, it has not been extensively explored for historical normalization of the given domain. Thus, this work gives a fresh perspective on established machine learning applications in historical research that should encourage further discussion on normalization and resource efficiency in text and document classification.

\section{Historical Text Normalization}
Normalization of texts is often considered a preliminary step in text processing, encompassing a variety of measures such as adjusting casing, removing stopwords or tags, changing punctuation, correcting spelling, and reducing repeated characters. These practices are well-established within the NLP community, although the specifics can vary from one study to another~\cite{Uysal_Gunal_2014}. Despite these traditions, normalization for historical texts presents unique challenges, primarily due to distinct linguistic variations and inconsistencies. Moreover, the representations of medieval documents in digital archives often vary significantly from their analog counterparts, a discrepancy exacerbated by the transmission from multiple sources and through documentary practices~\cite{Ehrmann_Hamdi_Pontes_Romanello_Doucet_2024}.

When it comes to diachronic languages, normalization takes on a more 'context-aware' approach. In the context of historical German texts, particularly those from the Middle High German (MHG) period, the necessity for and approach to normalization have been reinvigorated by recent advancements in data provision~\cite{Petran_Bollmann_Dipper_Klein_2016}, annotation~\cite{Vogeler_2019}, and analysis~\cite{Chinca_Young_2017}. This process involves converting historical texts into a modernized format, such as by updating or omitting characters, thereby making them more accessible for contemporary readers~\cite{Chiarcos_Kosmehl_Fäth_Sukhareva_2018,Sukhareva_2020}.\footnote{The importance of context-sensitive pre-processing has been raised in multiple studies which show that the selection and combination of normalization techniques significantly impacts the effectiveness of text classification; cf.~\cite{HaCohen-Kerner_Miller_Yigal_2020}.}

Notably, criticism has been raised against the concept of a universally accepted form of a normalized MHG, mainly due to the layers of normalization already introduced in past documentation and editing practices~\cite{Kragl_2015}. Furthermore, the differing priorities of diplomatists, who emphasize content, and linguists, who seek detailed transcriptions, complicate the quest for a single, definitive form. In this light, our work seeks to explore the possibility of classifying texts directly, which might allow us to bypass the need for extensive normalization altogether.

\section{Data}
The given charter data set has recently been prepared for a case study on MHG charters~\cite{Atzenhofer-Baumgartner_2023} and contains over 7,000 medieval charters derived from the charter platform \url{https://www.monasterium.net/mom/home} and is provided in a JSON serialization. The data contains archival identifiers, authenticated dates, original texts as per archival records, and normalized texts, as processed according to domain-specific procedures. Charters span from 1220 to 1390. They contain 385 words on average; the corpus size is about 2.8 million tokens.

The provided data set was processed rule-based. The rules were generated in an iterative concordance-informed workflow and entail mostly language simplification (e.g., 'ů' $\rightarrow$ 'u') and de-noising (e.g., '(sic) ' $\rightarrow$ ''). Around 17 \% of the documents have not been affected by the normalization procedure at all. Applying a standard gestalt pattern-matching algorithm to better understand the differences post-hoc, the transformation is estimated to have caused only around 68,000 character replacements and 15,000 deletions. In addition to this pre-processing, we only lower-case the texts.

Our study focuses on two distinct subsets: date ranges and archives. The dates provide temporal context for the dating task, while archives serve as indicators of regional text variation in the sense of geographical proxies. For the locating task, a subset is used which does not hold documents that are transmitted only through editions and which have no clear geographical reference. After removing these, the locating task is done with roughly half the data set.

\section{Methodology}
Throughout the years, a multitude of techniques and models have been used for text classification~\cite{Li_Peng_Li_Xia_Yang_Sun_Yu_He_2022}. 
Following recent trends that are empowering low-resource approaches against the backdrop of LLMs (e.g., compressors~\cite{Jiang_Yang_Tsirlin_Tang_Dai_Lin_2023} in the tradition of~\cite{Keogh_Lonardi_Ratanamahatana_2004} and~\cite{Cilibrasi_Vitanyi_2005}), we test three traditional classifiers and three more recent ones.
The former includes logistic regression (LR), naive bayes (NB), and support vector machines (SVM)~\cite{Cortes_Vapnik_1995}. The latter includes extreme gradient boosting (XGBoost~\cite{Chen_Guestrin_2016}), and two transformer-based models: RoBERTa~\cite{Liu_Ott_Goyal_Du_Joshi_Chen_Levy_Lewis_Zettlemoyer_Stoyanov_2019} and DeBERTa~\cite{He_Liu_Gao_Chen_2021}. We put a special focus on SVM by adding an enhanced version with minimally tuned hyper-parameters, given its long-standing competitiveness as well as robustness for the given task~\cite{Wahba_Madhavji_Steinbacher_2023} and domain register~\cite{Clavié_Alphonsus_2021}. We include transformers since their decent performance has been proven for historical (legal) texts ~\cite{Torres_Aguilar_2022}, but we do not incorporate  multilingual versions.\footnote{For this work, we prioritize lower-resource methods over LLMs for pragmatic reasons. Classical models require less computational power and are more suited to the smaller, specialized data sets typical of historical documents. They can offer a cost-effective alternative without sacrificing performance, while also providing a better and more readily available model explainability.}

As extracted features, TF-IDF is used. We also conduct 5-fold cross validations for all classifiers. For this reason, classes with too few occurrences for this setup are discarded, leading to 13 classes for decades (i.e., 1270 to 1400), and 20 classes for locations (e.g., archives in three different German-speaking countries). Regarding classes, we identify a decent balance for the decades of the 14\textsuperscript{th} century. More considerable class imbalance is seen in the archival classes.

\section{Results}
\begin{table}[!ht]
\centering
\caption{Dating metrics; non-normalized / normalized score. Best score is bold.}
\begin{tabular}{|l|cccc|}
\hline
Classifier & Accuracy & Precision & Recall & F1-Score \\ 
\hline
LR & 52.08 / 50.98 & 50.49 / 48.21 & 42.37 / 40.94 & 43.46 / 41.83 \\ 
\hline
NB & 30.42 / 29.88 & 33.08 / 27.96 & 18.97 / 18.46 & 15.20 / 14.35 \\ 
\hline
SVM & 57.56 / 57.51 & 56.12 / 56.50 & 49.23 / 48.70 & 50.20 / 49.75 \\ 
\hline 
SVM+ & 63.38 / 62.94 & \textbf{63.32} / 62.70 & \textbf{56.99} / 56.33 & \textbf{58.35} / 57.70 \\ 
\hline
XGBoost & \textbf{67.01} / 66.56 & 62.43 / 59.81 & 56.31 / 54.99 & 57.57 / 56.08 \\ 
\hline
RoBERTa & 34.10 / 32.89 & 28.84 / 27.16 & 29.22 / 28.02 & 27.25 / 24.74 \\ 
\hline
DeBERTa & 36.32 / 35.08 & 31.45 / 28.66 & 31.50 / 29.33 & 29.25 / 26.52 \\ 
\hline
\end{tabular}
\label{tab:dating-charters}
\end{table}

\begin{table}[!ht]
\centering
\caption{Locating metrics; non-normalized / normalized score. Best score is bold.}
\begin{tabular}{|l|cccc|}
\hline
Classifier & Accuracy & Precision & Recall & F1-Score \\ 
\hline
LR & 76.34 / 75.83 & 75.62 / 75.00 & 64.18 / 63.64 & 66.07 / 65.55 \\ 
\hline
NB & 37.04 / 36.24 & 36.17 / 35.08 & 21.16 / 20.60 & 19.44 / 18.77 \\ 
\hline
SVM & 81.88 / 82.01 & 85.35 / 85.21 & 74.42 / 74.72 & 76.95 / 77.18 \\ 
\hline
SVM+ & 85.35 / \textbf{85.48} & \textbf{87.63} / 87.50 & 79.37 / 79.54 & 81.62 / 81.70 \\ 
\hline
XGBoost & 85.22 / \textbf{85.48} & 86.05 / 87.32 & 82.32 / \textbf{83.61} & 83.27 / \textbf{84.81} \\ 
\hline
RoBERTa & 51.94 / 36.97 & 40.98 / 28.76 & 40.83 / 30.70 & 36.18 / 25.23 \\ 
\hline
DeBERTa & 53.79 / 47.93 & 43.42 / 34.30 & 43.28 / 38.28 & 39.04 / 32.71 \\ 
\hline
\end{tabular}
\label{tab:locating-charters}
\end{table}

As can be taken from Tables \ref{tab:dating-charters} and \ref{tab:locating-charters}, the present historical text normalization procedures appear to have a nuanced effect on the performance of classifiers. For the dating task, normalization generally reduces classification scores, suggesting that the original, non-normalized text holds unique features beneficial for this specific task. Conversely, for the locating task, normalization slightly improves or maintains performance for most classifiers, indicating that the process may help to highlight geographical indicators or standardize location-specific language variations.

Among the evaluated classifiers, SVM and XGBoost consistently and greatly outperform others in both dating and locating tasks. The enhanced version of SVM (SVM+) shows particularly strong results, achieving the highest scores in several metrics across both tasks. This demonstrates the value of fine-tuning traditional machine learning models for text classification tasks.

Interestingly, transformer-based models show a reduced performance across both tasks when text normalization is applied. This suggests that the inherent sub-word tokenization  may be less compatible with the modifications introduced through normalization which impacts their ability to capture traces of historical and geographical features that are present in the documents.

Furthermore, the results indicate a clear difference in difficulty between the dating and locating tasks, with classifiers generally achieving higher scores in the locating task. Again, we assume a presence of more distinctive language patterns associated with specific locations or archives, as opposed to the more subtle temporal cues required for accurate dating.

\section{Conclusion}

In conclusion, our investigation into the classification of medieval charters reveals an ambivalence, where the benefits of the given historical text normalization steps vary across tasks. Specifically, while normalization aids in the identification of locations, it may impair accuracy in dating documents. This underscores the nuanced impact of pre-processing and advises a cautious, task-specific approach to normalization. 

Notably, our findings highlight the superior performance of traditional classifiers like SVM and XGBoost over the transformer-based models for this application. While we acknowledge the evidence for the potential of 'historical transformers'~\cite{Manjavacas_Fonteyn_2022,Minaee_Kalchbrenner_Cambria_Nikzad_Chenaghlu_Gao_2022}, our results point to the importance of model selection in low-resource settings. We argue that leveraging the latest technologies must be balanced with the effectiveness and simplicity of existing learning methods.

Moreover, our research questions the necessity of extensive text normalization for classification and retrieval purposes, since we find that minimal or no pre-processing coupled with traditional classifiers yields effective results. We suggest a strategic shift either away from or towards customized pre-processing efforts that focus on preserving features that are potentially crucial for classification. 

Our study's implications gain importance when considering the wider context: even the largest charter databases hold only a portion of all available documents, leaving millions more in archives yet to be digitized and studied. We stress the need for improved retrieval methods: by refining them, we empower scholars to open the door to yet untapped historical knowledge.
%
%
%
\bibliographystyle{splncs04}
\bibliography{TPDL.bib}
\end{document}